\documentclass{article}
\usepackage{spconf,amsmath,graphicx}
\usepackage[utf8]{inputenc} 
\usepackage[T1]{fontenc}    
\usepackage{hyperref}       
\usepackage{url}            
\usepackage{booktabs}       
\usepackage{amsfonts}       
\usepackage{nicefrac}       
\usepackage{microtype}      
\usepackage{cite}
\usepackage{graphicx,psfrag,epsf}
\usepackage{enumerate}
\usepackage{url} 
\usepackage{amssymb,amsfonts,mathtools}
\usepackage{bm}
\usepackage{color, colortbl}
\usepackage{algorithm}
\usepackage{algorithmic}
\usepackage{balance,cite}
\usepackage{multirow}
\usepackage{cuted}
\usepackage{booktabs}

\usepackage{amsthm}

\usepackage{multicol} 
\DeclareMathOperator*{\argmin}{arg\,min}
\DeclareMathOperator*{\argmax}{arg\,max}

\def\T{{ \mathrm{\scriptscriptstyle T} }}

\def\GB{\textsc{G}}
\def\Gbart{\textsc{Gbart}}
\def\G{\mathcal{G}} 

\def\de{\overset{\Delta}{=}}

\def\D{\mathcal{D}}
\def\mo{m}
\def\isG{\textsc{isG}}
\def\de{\overset{\Delta}{=}}

\def\group{g}

\title{VARIABLE GROUPING BASED BAYESIAN ADDITIVE REGRESSION TREE}
\name{Yuhao Su and Jie Ding}
\address{School of Mathematics and School of Statistics\\
University of Minnesota, Minneapolis, MN 55455, USA}
\begin{document}
%
\maketitle
\begin{abstract}
\vspace{-0.1cm}
Using ensemble methods for regression has been a large success in obtaining high-accuracy prediction. Examples are Bagging, Random forest,  Boosting, BART (Bayesian additive regression tree), and their variants. 
In this paper, we propose a new perspective named variable grouping to enhance the predictive performance. The main idea is to seek for potential grouping of variables in such way that there is no nonlinear interaction term between variables of different groups. Given a sum-of-learner model, each learner will only be responsible for one group of variables, which would be more efficient in modeling nonlinear interactions.
We propose a two-stage method named variable grouping based Bayesian additive regression tree (GBART) with a well-developed python package \href{https://pypi.org/project/gbart/}{gbart}. The first stage is to search for potential interactions and an appropriate grouping of variables. The second stage is to build a final model based on the discovered groups. Experiments on synthetic and real data show that the proposed method can perform significantly better than classical approaches.
\end{abstract}
%

\section{Introduction}
\label{sec_intro}
\vspace{-0.1cm}
We consider a general regression problem
$
Y = f(X)+\epsilon 
$
where the regression function $f$ is unknown, $Y$ is the variable of interest, $X = [X_1,\cdots,X_p]^\T$ are predictors, and $\epsilon$ denotes the white noise. 
This formulation is common in a variety of statistical analysis  and machine learning tasks (in particular supervised learning with continuously-valued $Y$). 
Various ensemble decision tree methods including Bagging~\cite{Bagging1996}, Gradient boosting~\cite{GradientBoosting2001}, Random forest (RF)~\cite{RF2001}, multivariate adaptive regression splines~\cite{friedman1991multivariate}, and Bayesian additive regression tree (BART)~\cite{BART2010} have been proposed and widely used to estimate the regression function. 
These methods use an ensemble of weak learners, typically nonparametric decision trees, to reduce variance as well as bias in approximating $f(\cdot)$. Despite their practical success, it remains a challenge to develop comprehensive theoretical tools for understanding those sum-of-learner ensemble methods. Some general results on rates of convergence for function estimation can be found in, e.g., \cite{yang1999minimax,biau2012analysis} and the references therein.

Another direction of research imposes more strict assumptions on the regression function. A common assumption is the additive model originally proposed in~\cite{stone1985additive}:
\begin{align}
  f(X) = \mu+\sum_{i=1}^{p} f_i(X_i) , \quad
  E\{ f_i(X_i) \}=0,  \label{eq5}
\end{align}
where $\mu$ is the expectation of $f(X)$, each $f_i$ is a scalar function, and the expectation is with respect to the distribution of $X_i$ (required to ensure identifiability). 
Under the additivity condition, the multivariate function can be estimated with better statistical efficiency and computational easiness. For example, each scalar function $f_i$ may be approximated with series expansions such as polynomials, wavelets, or B-splines \cite{wahba1990spline}. The nonlinear regression is then turned into a linear regression that involves an enlarged set of regression variables, and those variables usually require further selection~\cite{ding2018model}. 
Sparse additive models have also drawn much attention in recent works~\cite{ravikumar2009sparse,huang2010variable,babadi2010sparls,mileounis2010adaptive,bazerque2011group}. Additionally, a spline-based additive model for time series modeling  was proposed and studied~\cite{han2017slants}.
Extensions to second-order polynomial regression have been considered as well (e.g. in the recent work of \cite{ye2019high}). 
Nevertheless, it is not clear whether the additivity assumption is too strong. In fact, the additive model class is not invariant under linear transformations of $X$. 
Though techniques developed under the additive models could be theoretically used to model higher-order interactions  using multidimensional splines or polynomials, they are often unrealistic due to the curse of dimensionality. 

In this work, motivated by both literature of sum-of-learner ensemble tree methods and additive models, 
we proposed a new method for estimating the regression function and to enhance the predictive performance. 
The main idea is to amalgamate the generality of tree methods with the additive structure that typically leads to less estimation variance and faster convergence.
We will build our method based on BART, an ensemble tree method that naturally fits the Bayesian perspective and supports a Markov chain Monte Carlo (MCMC) based fast implementation.
In fact, BART and its extensions (e.g.~\cite{BART2010,linero2018bayesian1}) have already shown remarkable prediction accuracy compared with other popular decision tree methods~\cite{BART2010},
so we will also use BART as a benchmark algorithm to compare with. 
Our method, referred to as variable grouping based Bayesian additive regression tree (GBART), consists of two stages. 
The first stage is to search for potential interactions and an appropriate partition of variables. The second stage is to build the final model based on the discovered groups. 
Our main contributions are two folds. First, the proposed algorithm can naturally bring prior knowledge of additive structure into learning and prediction. 
Though ensemble tree is a classical supervised learning tool that can be traced back to 1960s~\cite{AID,TreeHistory}, existing methods can barely benefit from prior structural information of the unknown function $f$. For example, if it is known that $f(x_1,\ldots, x_4)$ can be written as $f(x_1,x_2) + f(x_3,x_4)$, the estimation accuracy 
of an appropriately design learning method could be faster than that of a method without such prior. 
Second, our proposed solution performs much better than the state-of-the-art approaches when a grouping pattern does exist. Our implementation and detailed documentation are available as a Python package at \href{https://pypi.org/project/gbart/}{gbart}. 
The remainder is structured as follows. We introduce some background on the variable grouping and BART in Section~\ref{sec_back}. Details of the proposed GBART are presented in Section~\ref{sec_GBART}.  Section~\ref{sec_exp} provides various experimental studies (on both synthetic and real datasets). We conclude the paper  in Section~\ref{sec_con}.

\section{Background} \label{sec_back}
\vspace{-0.1cm}

\subsection{Variable grouping}
\vspace{-0.1cm}

 We first introduce the concept of variable grouping. Suppose that an unknown function $f$ with two variables $x_1, x_2$ is written as 
\begin{align*}
f(x_1,x_2) = f_1(x_1)+f_2(x_2)+f_{1,2}(x_1,x_2)
\end{align*} 
for some functions $f_1,f_2,f_{1,2}$, where 
$f_{1,2}(x_1,x_2)$ cannot be expressed as an addition of  terms one of which involves only $x_1$ or $x_2$. Specifically, if $f_{1,2}(x_1,x_2)$ can be written as
$
g_{1,2}(x_1,x_2)+g_1(x_1)+g_2(x_2)
$, 
then both $g_1$ and $g_2$ must be constant functions. 
If $f_{1,2}(x_1,x_2)$ is not a constant function, it is called an interaction term, and variables $x_1, x_2$ interact within the same group. In general, a function can be expressed as a summation of non-constant functions of various orders of interactions (including first-order terms such as $f_1(x_1)$).
In particular, a function $f(x)$ can be written as
\begin{equation}
\label{eqn_reform}
f(x) = \sum_{s=1}^{m}f_s(\{x_i: i \in \group_s\})  
\end{equation}
where $\G=\{\group_s\}_{s=1,\ldots,m}$ is a partition of $\{1,\ldots,p\}$ such that
  1) a variable interacts with at least another one in the same group, and
  2) variables from different groups do not interact. 
A similar formulation in the linear regression case was studied in the context of group LASSO~\cite{yuan2006model}. 


\subsection{Bayesian additive regression tree}

We briefly review Random forest and its Bayesian counterpart BART. Let $T_b$ denote the $b$-th tree, $b=1,\ldots,B$. 
Each tree $T_b$ has an internal structure $S_b$ that contains information such as the selected splitting variables and splitting values on internal nodes. Each tree also has a set of values associated with each terminal node, which is denoted by $M$. 
Then both Random forest 
and BART can be written in the form of a sum-of-learner model
\begin{align*}
\widehat{f}(X) = \sum_{b=1}^{B}T_b(X;S_b,M_b).
\end{align*} 
The main difference between them is that BART is based on a generative model and Bayesian-based inference methodology. 
Compared with a single tree and other sum-of-learner models, BART can incorporate interactions and additive effects more easily due to the adaptation of tree structure in each weak learner~\cite{BART2010}. 
As we mentioned in the introduction, ensemble methods such as Random forest and BART do not exploit potential additive structures of the unknown regression function, which is a prior knowledge that has been commonly assumed in the context of additive models. This motivates us to propose a new methodology that attains both advantages of ensemble tree methods and additive structures.

\section{Variable grouping through Bayesian additive regression tree} \label{sec_GBART}
\vspace{-0.1cm}

\subsection{Solution with a known partition}
\vspace{-0.1cm}
\label{sec_soln}
In this section we provide a solution to solve the reformulated regression problem (\ref{eqn_reform}) with a known partition. 
Suppose that a partition $\G$ of $\{1,2,\ldots,p\}$ is given,  and $\D=\{ z_j\}_{j=1,\ldots,n}$ are the observations with $ z_j=[x_j,y_j]$, $x_j\in\mathbb{R}^{p},y_j\in\mathbb{R}$. To obtain a regression model from $\D$ and $\G$, we estimate $f(X)$ in (\ref{eqn_reform}) with a sum-of-learner model 
\begin{align*}
\sum_{b=1}^{B}T_b(X;S_b,M_b)	
\end{align*}
where each weak learner (tree) $T_b$ is a Bayesian regression tree used in BART. The main difference with BART during learning is the sampling of variables and trees. We  incorporate variable grouping information $\G$ by uniformly sampling $s$ from $\{1,\ldots,m\}$ and assigning $\group_s$ to each weak learner $T_b$, meaning that the available space of predictors is restricted to $\group_s$ only. In this way, each term $\widehat{f_s}(\{x_i : i \in \group_s\})$ in (\ref{eqn_reform}) is approximated by a sum of weak learners involving only $\{x_i : i \in \group_s\}$. Compared with a uniform sampling from all the variables for each weak learner, as was used in Random forest, BART, etc., our proposed sampling method concentrates computational resources on where is needed and thus the estimation is expected to be more accurate. This has been verified in our experimental studies. The pseudocode is outlined in Algorithm~\ref{algo:base}.

\begin{algorithm}[ht]
\small
\centering
\caption{Group based Bayesian trees: $\mo_{\G} \gets \GB(\D_{t}, \G)$}\label{algo:base}
\begin{algorithmic}[1]
\INPUT Observations $\D=\{ z_j\}_{j=1,\ldots,n}$, partition $\G$ of $\{1,\ldots,p\}$ 
\OUTPUT Trained model $\mo_{\G}$
\FOR{$b = 1 \to B$}
\STATE initialize the $b$-th Bayesian additive regression tree
\STATE uniformly draw a group from $\G$ and assign it to the $b$-th tree 
\STATE restrict the search space of variables to be the assigned group
\ENDFOR
\STATE Use back-fitting Markov chain Monte Carlo to update trees
\STATE  Return $\mo_{\G}$
\end{algorithmic}
\end{algorithm}

\subsection{GBART: a two-stage variable grouping based method}
\vspace{-0.1cm}

In estimating a general regression function $f(X)$,
it is helpful to consider its approximation in the form of (\ref{eqn_reform}).
As we mentioned in the last subsection, a higher predictive performance could be obtained by taking advantage of a variable grouping structure. 
However, unless such grouping structure is given as domain knowledge, we have to search for the most appropriate grouping in practice. We thus propose a two-stage method where the first stage is to search for proper variable grouping information given observations $\D$, and the second stage is to build the final model based on the partition $\G$ discovered in the first stage.
The second stage has been addressed in Subsection~\ref{sec_soln}. Next we focus on the first stage. 



We initialize with the trivial partition where all the variables form a group. In each round, we first search for a variable that is marginally most significant, measured by the increased estimation error when such variable forms an individual group. We then search for another variable that, when joining the group of the last significant variable, will greatly reduce the estimation error. After that, we separate out a group of two variables.
This procedure can be extended to discover higher order interactions (i.e. groups of more than two variables). However, due to space limit  we will focus on second-order nonlinear interactions in this conference paper.
A pseudocode of the first stage of GBART is presented in Algorithm~\ref{algo:Group}.

Details of the pseudocode are elaborated below. 
In \textit{line} 1, we split the data for evaluating validation errors in later stages. Though the validation data may be reused in later stages, it is theoretical valid as long as the size of the validation dataset is much larger than the number of validation times. Relevant theory will be included in a journal version of this work. 
In \textit{line} 2, $\G$ is a list storing the discovered groups (from a greedy search). 
In \textit{line} 3, various variable selection may be performed for dimension reduction purpose (see~\cite{ding2018model} and the references therein). 
In \textit{lines} 4-5, a sum-of-learner ensemble tree model is learned from the trivial grouping, and the validation error is recorded as a benchmark.  
In \textit{lines} 6-10, the marginally most significant variable $X_{i^*}$ is selected.
In \textit{lines} 11-15, we search another variable $X_{k^*}$ that is best grouped with $X_{i^*}$, as measured by the validation error. 
In \textit{lines} 16-17, we then update the grouping $\G$ and other temporary quantities.
In \textit{line} 18, a group was found in a round and the searching is  continuously applied to the remaining variables. The searching process is terminated once a stopping criterion is met. 
In \textit{lines} 19-20, we complete the variable grouping and return the discoverd partition $\G$.
The complete GBART algorithm is outlined in Algorithm~\ref{algo:GBART}.
\begin{algorithm}[tb]
\small
\centering
\caption{Interaction Search based Grouping (isG): $\G \gets \isG(\D)$}\label{algo:Group}
\begin{algorithmic}[1]
\INPUT Observations $\D=\{ z_j\}_{j=1,\ldots,n}$
\OUTPUT Variable grouping $\G$ 
\STATE Randomly split the dataset $\D$ into $\D_{t}$ for training and $\D_{v}$ for validation
\STATE Initialize grouping $\G = \emptyset$ 
\STATE (Optional) Apply variable screening techniques to reduce variables to $\{x_i: i \in I\}$ where $card(I) \leq p$
\STATE Train model $\mo_{\G_0} \gets \GB(\D_{t}, \G_0)$, where $\G_0$ denote the trivial partition $\{I\}$
\STATE Obtain the (mean squared) validation error $e_0$ using $\D_{v}$
\FOR{$i$ in $I$}
\STATE Let $\G_i \de \{i, I \setminus  i\}$
\STATE Obtain model $\mo_{\G_i} \gets \GB(\D_{t}, \G_i)$ and validation error $e_i$ 
\ENDFOR
\STATE Let $i^* = \argmax_{i \in I} e_i$ 

\FOR{$k$ in $I$ and $k \neq i^*$}
\STATE Let $\G_k \de \{\{i^*, k\}, I \setminus  \{i^*, k\} \}$
\STATE Obtain $\mo_{\G_k} \gets \GB(\D_{t}, \G_k)$ and validation error $e'_k$
\STATE Let $k^* = \argmin_{k \in I, k \neq i^*} e'_k$
\ENDFOR
\STATE Let $g \de \{i^*, k^*\}$
\STATE Update  
			$I \gets I \setminus g$, and
			$\G.append(g)$ 
\STATE Repeat \textit{lines} 4-17 until $e_0 < e'_{k^*}$ or $card(I) \leq 1$
\STATE $\G.append(I)$ 
\STATE  Return $\G$
\end{algorithmic}
\end{algorithm}


\begin{algorithm}[ht]
\small
\centering
\caption{Variable Grouping based Bayesian Additive Regression Tree: $\mo_{\G} \gets \Gbart(\D)$}\label{algo:GBART}
\begin{algorithmic}[1]
\INPUT Observations $\D=\{ z_j\}_{j=1,\ldots,n}$, partition $\G$ of $\{1,\ldots,p\}$ 
\OUTPUT Trained model $\mo_{\G}$
\STATE Run Algorithm~\ref{algo:Group} to obtain $\G \gets \isG(\D)$
\STATE Run Algorithm~\ref{algo:base} to obtain $\mo_{\G} \gets \GB(\D, \G)$
\STATE  Return $\mo_{\G}$
\end{algorithmic}
\end{algorithm}

\section{Experimental Studies} \label{sec_exp}
\vspace{-0.1cm}

In this section we introduce our experimental results comparing GBART against BART and Random forest on both synthetic and real datasets. We start by introducing the experimental setting.
Our comparison is based on our developed Python package \href{https://pypi.org/project/gbart/}{gbart} that can perform both GBART and BART. The popular machine learning python package scikit-learn~\cite{sklearn} was used to implement Random forest. 
Throughout the experiments, 100 trees were used in the first stage of GBART to search for an appropriate variable grouping, and 200 trees were adopted in the second stage of GBART, as well as in BART and Random Forest. 
Performance is evaluated using the mean squared error (MSE) based on  five-fold cross-validation. To show the significance of differences in performance comparison, each experiment has been independently replicated 30 times to obtain the standard errors. 
We reported 12 synthetic datasets of various nature. The first 11 datasets were generated according to commonly considered nonlinear functional relations including multiplication, sum-of-squares, trigonometric functions. And we used the well-known Friedman dataset~\cite{friedman1991,Bagging1996} to be our last generated dataset. Details of generating functions and dimensions in 12 datasets are presented in Table~\ref{tab_setting}. 

\begin{table}[tb]
\centering
\small 
\begin{tabular}{p{0.45cm}p{0.45cm}p{6.5cm}}
\toprule
No.  & $p$ & Data Generating Function \\ \midrule
1   & $6$   & $Y = (x_1+x_2)^2+(x_3+x_4)^2+(x_5+x_6)^2+\epsilon$\\
2   & $6$     & $Y = x_1x_2+x_3x_4+x_5x_6+\epsilon$\\
3   & $6$     & $Y = x_1x_2+x_3+x_4+x_5+\epsilon$\\
4   & $6$     & $Y = x_1x_2+x_3x_4+x_5+x_6+\epsilon$\\
5   & $6$     & $Y = sinx_1sinx_2+(x_3+x_4)^2+(x_5+x_6)^2+\epsilon$\\
6   & $20$    & $Y = 5(x_1+x_2)^2+(x_3+x_4)^2+0.2(x_5+x_6)^2+0.04(x_7^2+x_8^2+\dots+x_{20}^2)+\epsilon$\\
7   & $20$    & $Y = 5x_1x_2+x_3x_4+0.2x_5x_6+0.04(x_7+x_8+\dots+x_{20})+\epsilon $\\
8   & $20$    & $Y = 5\sin x_1\sin x_2+(x_3+x_4)^2+0.2(x_5+x_6)^2+0.04(x_7+\dots+x_{20})+\epsilon $\\
9   & $20$   & $Y = 5\sin x_1\sin x_2+x_3x_4+0.2x_5x_6+0.04(x_7+\dots+x_{20})+\epsilon$\\
10  & $20$   & $Y = 5(x_1+x_2)^2+(x3+x4)^2+0.2(x_5+x_6)^2+\epsilon$\\
11  & $20$    & $Y = 5x_1x_2+x_3x_4+0.2x_5x_6+\epsilon$\\
12  & $7$   & $Y = 10\sin\pi x_1x_2+20(x_3-0.5)^2+10x_4+5x_5+\epsilon$\\\bottomrule
\end{tabular}
\caption{Twelve data generating models used for synthetic data experiments. The 12 case indices are corresponding to those in Table~\ref{tab_synthetic_result}.}
\label{tab_setting}
\end{table}
In the first 6 datasets, $x_1,\ldots,x_6$ were drawn from a multivariate normal distribution with a unit mean vector and identity covariance matrix. The additional 14 variables $x_7,\dots,x_{20}$ (whenever applied) are uniformly drawn from the interval $[0,1]$. Finally, the white noise $\epsilon\sim \mathcal{N}(0,0.5^2)$ was added to all datasets. In Friedman dataset, all variables are uniformly draw from $[0,1]$, with noise $\epsilon\sim \mathcal{N}(0,1)$. 
In the real data experiment, we used the concrete slump test dataset~\cite{slumptest} to evaluate our method. 
Concrete is a mixture of many different materials, and water plays a bonding role. It is known that the performance in concrete slump test is mainly attributed to a group of variables containing water and other related variables~\cite{slumptest}. 
Since there are three output variables, three different experiments were performed on the slump test data.

\begin{table}[tb]
\centering
\small 
\begin{tabular}{@{}llll@{}}
\toprule
No. & GBART                & BART         & RF            \\ \midrule
1  & \textbf{9.03 (0.60)}  & 10.28 (0.55)  & 23.61 (1.94)   \\
2  & \textbf{0.81 (0.05)}  & 1.05 (0.05)   & 2.66 (0.13)     \\
3  & \textbf{0.57 (0.02)}  & 0.73 (0.04)   & 1.34 (0.05)    \\
4  & \textbf{1.09 (0.06)}  & 1.30 (0.07)   & 2.07 (0.09)    \\
5  & \textbf{4.48 (0.41)}  & 6.46 (0.3)    & 12.71 (0.94)   \\
6  & \textbf{80.99 (9.86)} & 114.37 (6.97) & 232.57 (19.27) \\
7  & \textbf{9.88 (1.06)}  & 17.15 (1.14)  & 20.43 (1.22)   \\
8  & \textbf{7.83 (0.58)}  & 8.25 (0.41)   & 13.11 (0.73)   \\
9  & \textbf{2.67 (0.19)}  & 3.62 (0.13)   & 3.14 (0.11)    \\
10 & \textbf{75.9 (6.67)}  & 121.15 (9.36) & 254.8 (23.03) \\
11 & \textbf{9.79 (1.31)}  & 18.15 (1.21)  & 21.15 (1.38)    \\
12 & \textbf{1.98 (0.07)}  & 2.8 (0.08)    & 4.54 (0.12)    \\ \bottomrule
\end{tabular}
\caption{Performance comparison between the proposed GBART, BART, and RF on 12 different synthetic datasets, in terms of mean squared error. Standard errors from 30 independent replications are also reported (in the parenthesis). The optimal values are highlighted in bold.}
\label{tab_synthetic_result}
\end{table}

\begin{table}[tb]
\centering
\small 
\begin{tabular}{@{}llll@{}}
\toprule
Name           & GBART                 & BART         & RF           \\ \midrule
slump\_test1 & \textbf{5.34 (0.42)}   & 6.04  (0.45)   & 14.97 (0.9)  \\
slump\_test2 & \textbf{49.48  (1.28)}  & 53.44 (1.33)  & 51.29 (3.42)   \\
slump\_test3 & \textbf{162.86  (2.33)} & 179.39 (2.45) & 178.95 (9.32) \\ \bottomrule
\end{tabular}
\caption{Performance comparison between the proposed GBART, BART, and RF on three different \textit{slump tests}, in terms of mean squared error. Standard errors approximated from 30 re-samplings are also reported (in the parenthesis). The optimal values are highlighted in bold.}
\label{tab_real_result}
\end{table}
Table~\ref{tab_synthetic_result} and Table~\ref{tab_real_result} show that the proposed GBART method performs significantly better than other approaches. These results support our idea that by exploiting nonlinear interactions, our proposed method can be more statistically efficiently in modeling nonlinear interactions. 

\section{Conclusions} \label{sec_con}
\vspace{-0.0cm}

We proposed a new Bayesian additive tree method based on variable grouping to enhance the predictive performance in supervised learning. 
Our method was motivated by ensemble tree methods which typically ignore variable grouping structures  inherent to the unknown regression function.
Experimental results show a much better performance compared with classical methods when such variable grouping exist. 

An interesting future work is to apply the method to time series modeling.
Another future work is to extend the method from regression to classification tasks. 
\vfill\pagebreak
\clearpage

\balance
\bibliographystyle{IEEEbib}
\bibliography{gbart,references_com,J}

\end{document}